%% file: template.tex
\documentclass{article}
\usepackage{arxiv}
\usepackage{hyperref}
\hypersetup{
  pdfinfo={
    Title={Retargeting video with an end-to-end framework},
  }
}

\usepackage[utf8]{inputenc} 
\usepackage[T1]{fontenc}    
\usepackage{hyperref}       
\usepackage{url}            
\usepackage{booktabs}       
\usepackage{amsfonts}       
\usepackage{nicefrac}       
\usepackage{microtype}      
\usepackage{lipsum}
\usepackage{graphicx}
\graphicspath{ {./images/} }
\usepackage{amsmath}
\usepackage[numbers]{natbib}
\usepackage{dirtytalk}
\usepackage{subfig}
\usepackage{xcolor,soul,framed} 
\usepackage{enumitem}
\usepackage{algorithm} 
\usepackage{wrapfig}

\usepackage{array}
\usepackage{float}
\usepackage{mdwmath}
\usepackage{amsmath}
\usepackage{algpseudocode}
\usepackage{mdwtab}
\usepackage{eqparbox}
\usepackage{url}
\usepackage{csvsimple}
\usepackage{amssymb}
\usepackage{multicol}
\usepackage{multirow, tabularx}
\usepackage{adjustbox}
\usepackage{makecell}
\usepackage{siunitx}

\title{Retargeting video with an end-to-end framework}

\author{
 Thi-Ngoc-Hanh Le\\
 National Cheng-Kung University\\
  Taiwan\\
  \texttt{ngochanh.le1987@gmail.com}
   \And
 HuiGuang Huang \\
 National Cheng-Kung University\\
  Taiwan\\
  \texttt{604300468@qq.com}
  \And
 Yi-Ru Chen \\
 National Cheng-Kung University\\
  Taiwan\\
  \texttt{y2133256@gmail.com}
  \And
  Tong-Yee Lee*\\  
  National Cheng-Kung University\\
  Taiwan\\
  \texttt{tonylee@mail.ncku.edu.tw}
}

\begin{document}
\maketitle

\begin{abstract}
Video holds significance in computer graphics applications. Because of the heterogeneous of digital devices, retargeting videos becomes an essential function to enhance user viewing experience in such applications. In the research of video retargeting, preserving the relevant visual content in videos, avoiding flicking, and processing time are the vital challenges. Extending image retargeting techniques to the video domain is challenging due to the high running time. Prior work of video retargeting mainly utilizes time-consuming preprocessing to analyze frames. Plus, being tolerant of different video content, avoiding important objects from shrinking, and the ability to play with arbitrary ratios are the limitations that need to be resolved in these systems requiring investigation. In this paper, we present an end-to-end RETVI method to retarget videos to arbitrary aspect ratios. We eliminate the computational bottleneck in the conventional approaches by designing RETVI with two modules, content feature analyzer (CFA) and adaptive deforming estimator (ADE). The extensive experiments and evaluations show that our system outperforms previous work in quality and running time.

\end{abstract}

\section{Introduction}
Video is a widely used media form that holds significant importance in computer graphics applications due to its ability to convey motion, simulate reality, and engage viewers. Because of the heterogeneous of digital devices, adapting videos to different display size, resolution, or aspect ratios (referred to as \say{\textit{Video Retargeting}}) has become an essential function in these applications. For example, video retargeting can be employed to (1) adapt video content in real-time based on user actions or analytical queries \textcolor{black}{\citep{Adobe-After-Effects}}; (2) ensure that the video content is visually pleasing, legible, and seamlessly integrated with the augment reality environment \textcolor{black}{\citep{ARCore}}; or (3) dynamically adjusts the content of virtual reality (VR) videos to match the specific characteristics of different VR headsets \textcolor{black}{\citep{Unity}}. Additionally, resizing videos has become increasingly popular with the advent of smartphones equipped with video capabilities and the rise of social media platforms. Whether you choose a standard aspect ratio for content to be streamed on a laptop or a cell phone, it is crucial to ensure that the video presents clearly without unsightly cropping, providing the best experience for your audience. By leveraging video retargeting in these applications, it could ultimately enhance users overall viewing experience. To drive a video to a target size, conventional video retargeting methods require a time-consuming and expensive process to analyze input videos prior to resizing. In contrast, we have developed a deep learning-based framework allows for retargeting videos to arbitrary aspect ratios in an end-to-end manner. Our approach eliminates the computational bottleneck present in conventional methods while delivering higher quality results. 

Researchers explore the problems of video retargeting with various approaches, including conventional and state-of-the-art (SOTA) techniques. The retargeting image could be the background research of video domain. Several attempts have resolved the image retargeting's problem. Each method usually consists of two main steps: (1) determining the importance of image pixels by extracting an importance map and (2) performing an image retargeting operator to obtain the retargeting image \citep{asheghi2022comprehensive}. Deep learning-based techniques can produce good visual results. However, it requires equipment with high computation power and comprehensive datasets. We can see a comparison on the running time of the most recent SOTA image retargeting systems in \textcolor{black}{Fig. \ref{compare_timing}}. Extending a deep learning-based image retargeting technique to the video is tough to operate.

\input{Timing_images}

Unlike retargeting images, working on the video domain is more challenging. Along with preserving important content after resizing sessions, producing temporally coherent retargeting videos is also vital to judging the performance of such a video retargeting system. \textcolor{black}{Researchers have explored this research domain in various ways \citep{wang2014deformable, liu2016towards, zhang2008shrinkability, lin2013content, lee2020object}}. However, they share an identical workflow as those in the image domain. They need to utilize some off-the-shelf content analysis, i.e., saliency map and segmentation, to pre-process video frames in advance. This approach leads to two \textcolor{black}{significant} downfalls. First, the quality of resized videos depends on the performance of these auxiliary functions. This may not only downgrade the capability of the proposed method but also prevent the method from playing with diverse video content. Second, analyzing video frames is more expensive than single images. It may take minutes to process on a single frame. Processing time is an important factor when retargeting is used in services like video on demand or live streams, especially when the ability to retarget videos with high resolutions in real-time is required \citep{kiess2018survey}. Therefore, using such a time-consuming system to retarget video in the high demand of the digital age could be impractical.

To this end, the challenges in video resizing are not only retaining important content in such temporal coherent results but also the demand for a scheme that can adapt to diverse video content, tolerate various aspect ratios, and be fast in running time. Motivated by this observation, we propose a novel end-to-end framework, abbreviated as \textbf{RETVI}, to address the above challenges. We aim to retarget videos to arbitrary aspect ratios without using hand-craft pre-processing or image/video retargeting annotations. Our designed framework attempts to capture the critical information of video and preserve them after retargeting session without distortion and shrinking artifacts. Our essential contribution is reducing the running time cost to process video retargeting in an end-to-end manner. To achieve these, we design our framework consisting of a content feature analyzer (CFA) and an adaptive deforming estimator (ADE). The CFA is responsible for learning the content information of the input frames. Analyzed features serve as guidance for deforming the frame to a target size. In this manner, we propose a neural network, which can be considered an alternative to the time-consuming pre-processing in prior work. With the knowledge learned from CFA and a given resizing ratio, our ADE module tries to teach the network how to deform the input frame with minimum distortion. We test the performance of our proposed method through various videos. The experimental results and evaluations demonstrate that our system outperforms previous works in quality and running time. 

In summary, our main contributions are as follows:
\begin{itemize}
    \item We investigate a novel end-to-end video retargeting based on high-level representation of video frames.
    \item We design a network that can effectively demonstrate the important content of the video, which performs better the expensive computation of the preprocessing in prior video resizing systems.
    \item Our method can adapt to various video contents and arbitrary aspect ratios.
    \item End-to-end and fast running time enable our system to be potentially embedded in other services or applications.
\end{itemize}

\section{Related work}
\vspace{-0.25cm}
\subsection{Image Retargeting}
\textcolor{black}{Image Retargeting has received much attention during the last decade. Researchers approach this domain with different techniques, from conventional to state-of-the-art. Most of the classical image resizing algorithms \citep{3_avidan2007seam, liu2003automatic, hashemzadeh2019content, 1_lin2012patch, jin2010nonhomogeneous, dong2015image} share an identical workflow with two steps. In the first step, an importance map is generated via visual attention analysis methods, like saliency detection. The main goal is to preserve the regions with high importance as well as possible. In the second step, an operator retargets the image \citep{kiess2018survey}. Cropping, scaling, warping, browsing, seam carving, or combining them are the typical operators selected and used in these resizing algorithms. Most recently, \citet{kim2018quad} investigated a novel approach, a grid encoding model for image retargeting, which takes each horizontal/vertical distance between two adjacent vertices as an optimization variable. The results demonstrate that their method consistently outperforms previous methods on qualitative and quantitative perspectives \citep{kim2018quad}.}

\textcolor{black}{With the revolution of deep learning technologies, several works have applied these new techniques to resolve the problems in image retargeting \citep{zhou2020weakly, song2018carvingnet, tan2019cycle, liu2018composing, cho2017weakly, lin2019deepir}. CarvingNet \citep{song2018carvingnet} uses an encoder-decoder CNN to develop an importance map based on a learning model. WSSDCNN \citep{cho2017weakly}, a network learns the image content via an attention map, guiding pixel-wise mapping during retargeting. Cycle-IR \citep{tan2019cycle} solves the problem of image retargeting through unsupervised learning. Conceptually, they build the network based on reverse mapping from the retargeted images to the given input images. Formulating the multi-operator retargeting upon reinforcement learning technique is a novel approach presented in SAMIR \citep{zhou2020weakly}. This approach can produce results with lower computational costs. Nevertheless, the running time is relatively high, and the results still suffer from cropping seriously.}

\vspace{-0.25cm}
\subsection{Video Retargeting}
\vspace{-0.25cm}
\textcolor{black}{Preserving the relevant visual content in videos while avoiding flicking is the most crucial challenge of video retargeting \citep{kiess2018survey}. Using image retargeting techniques on individual video frames does not provide satisfactory results as they might change entirely different areas in adjacent frames \citep{kopf2009fscav}. Video retargeting becomes an exciting research topic recently when the explosive growth of social platforms and digital devices demands the videos to be resized to display them nicely. }

The early attempts to retarget videos use a straightforward technique, cropping \citep{liu2016towards, cheng2006video, zhang2013compressed}. Later, researchers investigate more algorithms to resolve the problems in this research domain. Extending the typical techniques in image retargeting to video is commonly used by researchers, particularly seam carving and warping operators. Extending seam carving to video retargeting might lead to high processing time. Therefore, researchers in \citep{furuta2016fast, grundmann2010discontinuous, wang2014deformable} focus on speeding up the processing time and saving memory space when extending this technique to videos. Applying warping to retargeting videos may cause undesirable artifacts. It is because of a temporal motion of an image region in one direction followed by motion in the opposite direction. Therefore, warping-based research applied to videos focuses on image stability and run-time performance \citep{kiess2018survey}. 
\citet{gallea2014physical} use a 2D grid for image retargeting and add a third dimension to handle changes over time. The authors in \citep{li2014spatiotemporal, yan2013effective} extend the axis-aligned image retargeting to video by computing the deformation for selected keyframes and interpolating the other frames.
A contrasting approach to these extensions is introduced by \citet{lin2013content}. The authors use a uniform grid mesh for the warping in this work. They focus on preserving the important objects in a video while warping the non-important regions in a way similar to linear scaling \citep{kiess2018survey}. The most recent work in this research domain is introduced by \citet{lee2020object}, which uses a deep neural network for video retargeting. Their concept uses object detection to allocate the bounding boxes of the main objects in a video. Then, the remaining background areas are resized to preserve the content of the bounding boxes without deformation. \textcolor{black}{Along with retargeting traditional images/videos, the research domain on stereo image/video retargeting also receives attention. The studies proposed in \citep{lin2013object, lin2015consistent, shao2017qoe, fan2021stereoscopic} demonstrate that analyzing the importance in input is an essential factor to gain good quality resized stereo images/videos.}

The sharp contrast between our framework and theirs \citep{wang2014deformable, zhang2008shrinkability, lin2013content, lee2020object} is that we develop an end-to-end system for retargeting videos to arbitrary aspect ratios. Our proposed method is faster than the conventional techniques, effectively preserving important information in videos after retargeting session without distortion or shrinking phenomenon, and tolerant to videos with dense content and multiple moving objects alike.
\input{Net.tex}

\section{Methodology}
\vspace{-0.25cm}
\subsection{System overview}
We propose a novel framework RETVI for retargeting videos to the arbitrary aspect ratio. Our approach is to retarget videos without hand-craft pre-processing or image/video retargeting annotations. We convert the video retargeting problem to an unsupervised learning problem of conditional transformation regression without explicitly computing a transformation. \textcolor{black}{As diagrammed in Fig. \ref{fig_Net}, a pair of video frames ($\mathcal{V}^o$, $\mathcal{V}^e$) is required in the training process, in which $\mathcal{V}^o$ is the input video frame and $\mathcal{V}^e$ is the extracted foreground of $\mathcal{V}^o$. In the inference phase, only the video with frames $\{\mathcal{V}^o\}$ is required as input in our end-to-end retargeting process}. In the following, in the context that we do not need to differentiate $\mathcal{V}^o$ and $\mathcal{V}^e$, we use $\mathcal{V}$ to denote such an input frame in our system. Specifically, we model our RETVI with two modules, a content feature analyzer (\textbf{CFA}) and an adaptive deforming estimator (\textbf{ADE}). \textcolor{black}{The specific procedure is outlined in Fig. \ref{fig_Net}(b)}.

As named, the CFA is responsible for analyzing the input content. It accepts video frames as the input. As a result, CFA converts the \textcolor{black}{input frames $\mathcal{V}$} to high-level representation. Conceptually, ADE receives feature representation \textcolor{black}{of $\mathcal{V}$} and a random resizing ratio as inputs, and tries to teach the network how to deform $\mathcal{V}$ with minimum distortion. Afterwards, we train the network with the weighted sum of four loss functions. Once trained, given a video and target size, our network can efficiently and effectively produce the retargeting result in an end-to-end manner. Instead of using an expensive pre-processing for visual attention analysis as in the aforementioned conventional retargeting systems (\textit{e.g.}, saliency map, segmentation, optical flow), we analyze the input frames by the CFA module. The features obtained from CFA are more robust to noise and occlusions than saliency maps or segmentations. This is because they rely on higher-level features, such as shape, texture, and context, which are less affected by noise and occlusions than low-level features such as color or intensity. Besides, CFA can be more computationally efficient and generalizable to new images once trained.

\subsection{Network architecture}
\vspace{-0.25cm}
\subsubsection{Content Feature Analyzer}
This module, dubbed as \textbf{CFA}, shoulders the task of learning the contextual information of frames during encoding them into latent space. This process can be considered as an alternative to the time-consuming pre-processing, \textit{i.e.,} saliency and segmentation, in those described above conventional resizing systems. \textcolor{black}{To define the important regions in frames, such a well-known object detection also could be used. Nonetheless, detection focuses on detecting the bounding boxes surrounding the detected objects. Meanwhile, we aim at estimating at the pixel level to capture more semantic information in the frame rather than using the bounding boxes. Consequently, we structure CFA such that it can produce pixel-wise output of input frame from encoded features. }We design CFA with the inspiration from the U-Net model \citep{ronneberger2015u}, which has been a successful network in medical image segmentation. It is later varied to be used in other applications, such as natural image analysis, natural language processing, and image classification. U-Net is shaped in various structures depending on the goal of a particular application. Similarly, we design a variation of this standard model to meet the purpose mentioned earlier. The tweaks of our design are as follows.

Given a video \textcolor{black}{frame $\mathcal{V}$ in size $H \times W$,} \textcolor{black}{we generate} a pyramid of spatial features from coarse to fine granularities through a so-called E-Blocks, which is structured by \textit{Conv3$\times$3 $\rightarrow$ Normalization $\rightarrow$ Tanh}. Formally, the feature maps produced by an E-Block is formulated as: 
\begin{equation}
    \mathbf{E}_i = \mathcal{T}(\mathcal{B}(C^3(\mathbf{E}_{i-1}, \kappa_i))),
\end{equation} \textcolor{black}{here $C^3(.)$ performs the $3\times3$ convolution, $\mathcal{B}(.)$ represents the batch normalization, and $\mathcal{T}(.)$ indicates the Tanh activation function. $\mathbf{E}_i$ is a matrix in $\mathbb{R}^{H_i\times W_i \times \kappa}$ with $\kappa_i$ is the number of kernels used in the $i^{th}$ E-Block, $i=1 \dots 7$. The contrary path is a symmetric form of those obtained from E-Blocks}. It is worth noting that we deepen the number of layers to seven since we aim to detect both the more extensive features (\textit{i.e.,} the main objects in the foreground) and the more fine-grained features (\textit{i.e.,} the small objects in the background). \textcolor{black}{In the pure UNet and its variants, skip connections play a crucial role in facilitating information flow between different levels of the network, enabling better feature propagation and learning. They mostly use skip connections at every upsample layer. In contrast, we employ the skip connections half-way, \textit{i.e., }at layer 4 to layer 7, as features are with more fine granularity dwell deeper layers. This strategy could aid in discriminating objects  from the background regions or multiple objects distributed entire frame. It eventually boosts the capability of the decoder. } Specifically, we embed an \textit{ED-Gate} (\textbf{G}) at each skip connection to summarize the fine features and construct the decoded features. The feature maps at each symmetric layer are expressed as follows:
\begin{equation}
    \mathbf{D}_i = \begin{cases}
        \mathbf{G}(\mathbf{E}_i, \mathbf{D}_{i+1}) \text{  , if } 4 \leq i \leq 6 \\
        \Omega(\mathbf{D}_{i+1}) \text{ , if } 1 \leq i < 4
    \end{cases},
\end{equation} \textcolor{black}{here $\Omega(.)$ represents for a so-called D-Block, which is formulated as:}
\begin{equation}
    \Omega(\mathcal{F}^{out}) = \mathcal{R}(\mathcal{B}(C^T(\mathcal{F}^{in}))),
\end{equation} where $\mathcal{R}$ is the ReLu activation function, and $C^T(.)$ is the transposed convolution operator. In essence, the $C^T(.)$ operation forms the same connectivity as the normal convolution but in the backward direction. Moreover, the weights in $C^T$ are learnable, we accordingly do not need a pre-defined interpolation method. 

In the ED-Gate, given two feature maps $\mathcal{F}^{e}$ and $\mathcal{F}^d$, respectively from encoding and decoding layers, $\mathcal{F}^e$ is first fed to a CBR block, which is structured as \textcolor{black}{\textit{Conv3$\times$3 $\rightarrow$ Normalization $\rightarrow$ ReLu}.} It is then concatenated with $\mathcal{F}^d$ to result in the output feature maps at a certain skip connection. Theoretically, the output of an ED-Gate is expressed as:
\begin{equation}
    \mathcal{F}^g = \text{CBR}(\mathcal{F}^e) \boxtimes \mathcal{F}^d,
\end{equation} with $\boxtimes$ is the concatenation operation. 

We can obtain certain advantages with the above design. First, increasing the \textcolor{black}{depth of layers} can capture more complex features in the image, which can lead to improve the accuracy of understanding the image content. This is especially important for analyzing the image having complex textures. Second, the ED-Gate boosts that the information can flow more directly between the encoder and decoder. Third, this design can help improve the parameter efficiency of our model by allowing it to capture more information with fewer parameters. This can help prevent overfitting and improve the generalization performance of the model.

\subsubsection{Adaptive Deforming Estimator}
With the decoded feature map $\mathbf{D}_1$ obtained from the CFA, the question here is how to thread it to learn the appropriate deformation matrix. In other words, we need to define a function $f: \mathbb{R}^{h \times w\times k} \longrightarrow \mathbb{R}^{S_h \times S_w}$, where $h, w, k$ is the height, width, and kernel size of $\mathbf{D}_1$. The matrix $\mathbf{H} \in \mathbb{R}^{S_h \times S_w}$ is shaped in the same size with the input frame, \textit{i.e., $S_h \times S_w$}. Finding the mapping function $f$ is a game that several prior work has challenged with, such as, geometric warping with industrial style transfer \citep{yang2022industrial}, \textcolor{black}{video stabilization \citep{zhao2020pwstablenet}}. Optionally, depending on the goal of a certain application, the function $f$ is designed in different ways. In our current application, we aim to estimate a deformation to drive the input frame from a source size $\mathbf{S}$ to a target one $\mathbf{T}$, $\mathbf{S}^{S_h \times S_w} \longrightarrow \mathbf{T}^{T_h \times T_w}$, with minimum distortion and content awareness. For this goal, we design the Adaptive Deforming Estimator (\textbf{ADE}). In this stage, we consider the advantage of both pixel-based and grid-based strategies when formulating the function $f$. Pixel-based is simple and can be computationally efficient. However, it can lead to loss the detail and resolution, particularly when downsampling an image. Meanwhile, grid-based methods can preserve more detail and resolution in the resized frame than pixel-based methods. However, it can be more computationally expensive than pixel-based methods. To alleviate the burden of the trade-offs between computational efficiency and image quality, we devise an in-between strategy, which can use the advantage of each method.

\input{workflow}
\textcolor{black}{Now, we detail how we utilize feature maps $\mathbf{D}_1$ to deform frames. The workflow of this phase is outlined in Fig. \ref{fig_workflow}. First, we transform $\mathbf{D}_1$ to 2D-grid form via an activation function:}
\begin{equation}
    Q =  \frac{1-e^{-2a}}{1 + e^{-2a}},
\end{equation} here, $a \in \mathbf{D}_1$, a tensor with 16 channels; hence, $Q\in\mathbb{R}^{2 \times 16\times16}$. \textcolor{black}{Through equation (5), each element in the feature map $\mathbf{D}_1$ is returned values in range (-1, 1), which indicates the fine features of the input frame. To define the mapping value in $Q$ to pixel in $\mathcal{V}$, }we interpolate $Q$ to a regular $S_h\times S_w$ grid pixel coordinate: $\mathbf{E} = \nabla_b(Q)$, here $\nabla_b$ denotes the upsampling operator with bilinear mode. \textcolor{black}{The resultant dense map $\mathbf{E}$ represents for the information of the corresponding pixel in the input frame. As a result, each pixel in $\mathbf{E}$ is also a grid cell, and we assume each of them to be a seed.} A seed $s(c^s, e^s) \in \mathbf{E}$ has a coordinate $c^s(x, y)$ and \textcolor{black}{energy $e^s(x, y)$}. The energy of a seed demonstrates how important the pixel is. \textcolor{black}{A pixel belonging to important regions should have a smaller energy compared to those belonging to the reverse one. In other words, pixels are with higher energy will be more deformed than others}. We call $\mathbf{E}$ an \textit{energy map}, which is considered as the guidance for the later deforming estimation.

Given a target size $\mathbf{T}(T_h, T_w)$, to construct the resizing form (denoted as $\mathcal{R}$) of frame $\mathcal{V}$, instead of deforming $\mathcal{V}$ directly, we make a tweak to manipulate this process. \textcolor{black}{First, the energy of seeds in $\mathbf{E}$ is extracted and re-formulated to define a matrix $\mathbf{H}$, in which value at coordinate $(i, j)$ is a tuple with:
\begin{equation}
        \begin{cases}
        \mathbf{H}^{ij}_x = e^s_x \times \Re \\
        \mathbf{H}^{ij}_y = e^s_y \times \Re \\
    \end{cases}
\end{equation}
here $i = 0,\dots, S_h-1$; $j = 0, \dots, S_w-1$; and 
\begin{equation}
\Re =
\begin{cases}
        \big(1-r\big)^2 \text{,  if reducing} \\
        -\big(1-r\big)^2 \text{,  if enlarging}
\end{cases}
\end{equation}
where $r = T_w\big/S_w$, \textit{i.e., }the resizing ratio. To make our model be tolerant with arbitrary target sizes, $r$ is randomized in the range of $[0.25, \dots, 1.25]$ in our training}. The matrix $\mathbf{H}$ represents the seed's flow intensity when the current space is retargeted. Thereafter, we initialize an empty pseudo frame p-$\mathcal{V}$ in the same size with $\mathcal{V}$. Then, we partition $\mathcal{V}$ and p-$\mathcal{V}$ into a regular grid of size ($S_h\times S_w$). Consequently, we deform p-$\mathcal{V}$ using the matrix $\mathbf{H}$, resulting in a deformed grid d-$\mathcal{V}$. The coordinate of a cell in d-$\mathcal{V}$ is defined as: \textcolor{black}{($c^s_x + \mathbf{H}^{ij}_x, c^s_y + \mathbf{H}^{ij}_y$)}. \textcolor{black}{By applying different deformation weights to the  x-axis and y-axis separately, we can achieve non-linear changes in the shape of the resized frames.} Once \textcolor{black}{deforming} p-$\mathcal{V}$ to d-$\mathcal{V}$, we use d-$\mathcal{V}$ as a mapping to sample pixel values from $\mathcal{V}$ and generate resizing frame $\mathcal{R}$. The frame $\mathcal{R}$ is constructed by interpolating the pixel values according to d-$\mathcal{V}$. The matrix $\mathbf{H}$ is a learnable parameter, and $\mathcal{R}$ is optimized at each iteration during training by four loss functions. \textcolor{black}{The energy $e^s$ is contravariant with the pixel property at a particular seed. That is, the more important the pixel is, the smaller the energy is. }

Our above tweak serves the following benefits. First, the deformation is applied on the coordinate system itself, without considering the content of the frame. Hence, it results in a smooth and continuous transformation. Second, we have more control over the transformation of the coordinate space. This allows us to apply various aspect ratios on the deformation. Besides, by formulating $\Re$ during the training, the model can learn to adapt to various resizing ratios under the control of loss functions. As a result, we do not need to re-train the network whenever it plays with a new target size. Also, such a time-consuming pre-processing mentioned above is eliminated and videos are retargeted in an end-to-end process.

\subsection{Loss function}
Let the generator network be denoted by $f$ parameterized by \textcolor{black}{weights $\mu$}, it transforms an input \textcolor{black}{frame $\mathcal{V} \in \mathbb{R}^{H \times W \times C}$} into an output deformation matrix $\mathbf{H}$ via the mapping $\mathbf{H} = f_{\mu}(\mathcal{V})$. Network learning adjusts the parameters $\mu$ through minimizing four loss functions $\mathcal{L}_{\textit{cri}}$, $\mathcal{L}_{\textit{glo}}$, $\mathcal{L}_{\textit{tem}}$, and $\mathcal{L}_{\textit{fid}}$. Each loss function computes a scalar value $\mathcal{L}(.)$ measuring the difference between the input frame and the retargeting frame corresponding to four loss functions. The network is trained to minimize a weighted combination of the loss function:
\begin{equation}
    \mu^* = \arg \min_{\mu} E_{\mathcal{V}}\big[\lambda_c\mathcal{L}_{cri} + \lambda_g\mathcal{L}_{glo} +
     \lambda_t\mathcal{L}_{tem} + \lambda_f \mathcal{L}_{fid}\big].
\end{equation}\textcolor{black}{The parameters $\lambda_c$, $\lambda_g$, $\lambda_t$ and $\lambda_f$ are all set to 1 in our training.} In the following, we delve in detail the figuration of each loss function. The ablated results on each component in our optimization are presented in later Sec.4.5.

\textbf{Critical region loss} ($\mathcal{L}_{\textit{cri}}$). In reconstructing the frame to a new ratio, the goal is to recover the impaired frame to match the pristine undistorted counterpart visually. Thus, we need to design the loss that would adhere to that goal. We are inspired by the perceptual loss \citep{34_johnson2016perceptual} to supervise feature changes. Perceptual loss is expressed in various forms depending on \textcolor{black}{specific application}, e.g., image style transfer, image restoration, image colorization. Specifically, in our current application, we formulate this loss to distinguish important objects in a frame and maintain their proportional shape between the ground truth foreground and the estimated foreground frame, written as:
\begin{equation}
    \mathcal{L}_{cri} = \sum_i \frac{1}{C_iH_iW_i} \parallel (\Gamma(\Phi_i(\mathcal{V}^e)) - \Gamma(\Phi_i(\mathcal{R}^e)) \parallel^2_2,
\end{equation}where $\Gamma$ is an off-the-shelf feature extractor. \textcolor{black}{In our experiment, we use VGG-19 as a feature extractor. However, other pre-trained models, such as ResNet, could result in equivalent effect.} $\Phi_i(.)$ is the feature maps in the $i^{th}$ layer of the corresponding input parameter $\mathcal{V}^e$ or $\mathcal{R}^e$. Here, $\mathcal{V}^e$ is annotated foreground of the input frame $\mathcal{V}^o$ and $\mathcal{R}^e$ is resized form of $\mathcal{V}^e$. With this loss function, the regions with \textcolor{black}{low energy} in the input frame can be well preserved.

\textbf{Global integrity loss} ($\mathcal{L}_{glo}$). 
In addition to the critical regions, we also consider the overall presentation of the output image. This objective function is particularly effective when the content is dense and distributed entire frame. To preserve the frame information as much as possible, making resized frames more harmonious and natural, we rely on the advantage of an image classification network (shortened by ICFNet). In a certain ICFNet, for instance, VGG or ResNet, the architecture consists of feature extraction blocks and ends with a fully connected layer, which is then used for producing predictions. Inspired by this, we formulate $\mathcal{L}_{glo}$ to define how the content of the input frame is preserved in the retargeted one. Let denote $x$ be the one-dimensional latent vector of an input frame $\mathcal{V}$ after the fully connected layer of the ICFNet. The probabilities in vector $x$ for all possible $N_c$ classes is expressed as:\textcolor{black}{
\begin{equation}
    P(x|N_c) = \frac{\text{exp}(x)}{\sum_{j=1}^{N_c} \text{exp}(x_j)}
\end{equation}}
 The objective of $\mathcal{L}_{glo}$ can be simplified as:
\begin{equation}
    \mathcal{L}_{glo} = \parallel P(x^o) - P(x^r) \parallel^2_2),
\end{equation}here $P$ is defined by Eq.(10); arguments $x^o$, $x^r$ are respectively the latent vectors of $\mathcal{V}^o$, $\mathcal{R}^o$ after feeding them to a ICFNet. In our experiment, the ICFNet we use is ResNet. Using VGG models may also result in equivalent effect. With Eq.(11), we can measure \textcolor{black}{the different degree of the pair ($x^o, x^r$)}, i.e., defining whether they are classified into the same classification. In this way, the content information can be compared without being limited by the size difference \textcolor{black}{between frames $\mathcal{V}^o$ and $\mathcal{R}^o$.}

\textbf{Temporal consistency loss} ($\mathcal{L}_{tem}$). 
Temporal consistency should be specifically considered in our video retargeting task. The temporal consistency loss is used for evaluating the coherence between adjacent resized frames. To obtain this coherency, several works in the video generation domain, such as video style transfer, video stabilization, video segmentation, \textit{etc.}, usually calculate the pixel correspondence between adjacent frames with optical flows. However, the time cost of optical flow estimation is expensive. It could affect the efficiency of the model. Besides, in our video retargeting application, the sizes of the input and retargeted frames are different, so finding the mapping optical flows in this context is impractical. Instead, we base on the knowledge of the correlations between the features of the adjacent frames to infer the temporal consistency degree in the retargeted videos. For any triplet of the adjacent frames with the respective supper script $t-1$, $t$, and $t+1$, the temporal consistency loss is formulated as:
\textcolor{black}{
\begin{equation}    
    \mathcal{L}_{tem} = \sum \parallel \gamma(\mathcal{V}^o_{t-1}, \mathcal{V}^o_t, \mathcal{V}^o_{t+1})
        - \gamma(\mathcal{R}^o_{t-1}, \mathcal{R}^o_t, \mathcal{R}^o_{t+1}) \parallel^2_2  
\end{equation}where $\gamma(.)$ is to calculate the correlation of features between adjacent frames. Assuming with a certain triplet of three adjacent frames ($\mathbf{F}_{t-1}, \mathbf{F}_t, \mathbf{F}_{t+1}$), $\gamma(.)$ is expressed as:}
\textcolor{black}{
\begin{equation}
    \gamma(.) = \frac{\sum\big(\Gamma(\Phi_i(\mathbf{F}_{t-1}))
    \times \Gamma(\Phi_i(\mathbf{F}_t))\big) \times \Gamma(\Phi_i(\mathbf{F}_{t+1}))\big)}{H_iW_iC_i}
\end{equation}}with $\Gamma(.)$ and $\Phi(.)$ are defined in the same way with Eq.(9); $i \in [1 \dots 4]$. Using the correlation of features between adjacent frames could be more appropriate to maintain temporal coherence than optical flow in video retargeting applications. Video retargeting involves changing the resolution of a video, which can cause distortion and loss of detail. To avoid these issues, it's important to maintain temporal coherence between frames, so that the motion and appearance of objects in the video remain consistent. Compared to other alternatives, such as optical flow, our formulation is simplicity, robustness, and \textcolor{black}{useful for our task when changing the size}.

\textbf{Fidelity loss} ($\mathcal{L}_{fid}$). 
Apart from minimizing geometry distortions, the quality of the resized frames is also a vital aspect we take into account in our current application. The matrix estimated by our network plays as \textcolor{black}{the weight for deforming frames}. However, the retargeted results might be blurry due to the relative coordinates we used to reconstruct the pixels in the ADE module. To alleviate this phenomenon, we propose the fidelity loss. Our idea in this manner comes from the concept of Generative Adversarial Networks (GANs). Unlike the discriminator in such a GAN-based network which tries to classify examples as either real (from the domain) or fake (generated), our formulation aims to discriminate the fidelity of retargeted frame to its corresponding input in terms of pixel color value. We note here that without $\mathcal{L}_{fid}$, our network alone is capable to estimate the \textcolor{black}{deformation} matrix efficiently. Besides, our model is not supervised by the annotated retargeting data, the optimization plays as auto-alignment manner. To this end, we propose this loss to boost the performance of our model in terms of \say{clean quality.}

A pair of frames ($\mathcal{V}^o, \mathcal{R}^o$) is taken as input in our formulation. They are first fed to a so-called \textit{Fid-Disc} to analyze them in feature space. Unlike other loss functions, $\mathcal{L}_{cri}$, $\mathcal{L}_{glo}$, $\mathcal{L}_{tem}$, which use an off-the-shelf feature extractor, we design a lightweight module to configure $\mathcal{L}_{fid}$. \textcolor{black}{The Fid-Disc contains 6 CBR blocks, as used in the ED-Gate, and 1 fully-connected hidden layer (64 channels)}. Accordingly, the fidelity loss is expressed as: 
\begin{equation}
    \mathcal{L}_{fid} = \frac{1}{2}\bigg(\xi(\mathcal{D}(\mathcal{V}^o), 1) + \xi(\mathcal{D}(\mathcal{R}^o), 0)\bigg),
\end{equation}
where 
\begin{equation}
    \xi(.) = -\frac{1}{n} \sum \big ((\tau[i]* log(\eta[i]))\\ 
    + (1-\tau[i]) * log(1-\eta[i])\big).
\end{equation}Here, $\mathcal{D}$ is the Fid-Disc; $\eta$ is the score output by $\mathcal{D}$ with the parameters $\mathcal{V}^o$ or $\mathcal{R}^o$; and $\tau$ is the score we want it to discriminate, i.e., $\tau \in \{0, 1\}$. The training goal of $\mathcal{D}$ is to distinguish the input frame from its resized frame by the ADE module. Our network tries to generate a high score (\textit{i.e.}, close to 1) from $\mathcal{D}$, so that the output of ADE module can be closer to the quality of the input frame.

\section{Experimental Results}
\vspace{-0.25cm}
\subsection{Implementation Details}
We implement our RETVI in Pytorch \citep{paszke2019pytorch}. All experiments were performed on a PC equipped with Intel Core i7-770 CPU, 16GB RAM and a GeForce RTX 2080 Ti GPU. We train RETVI using batch size 2 within 150 epochs. The training takes approximately 10 hours. For the loss function, we use Adam optimizer \citep{23_kingma2014adam} with a learning rate of $1 \times 10^{-3}$. 

In terms of training data, we train our RETVI on DAVIS dataset \citep{pont20172017}, which consists of 90 videos in size 854 $\times$ 480 with single and multiple moving objects. In the dataset, each video has a corresponding foreground annotation. We use it as $\mathcal{V}^e$ and the original video as $\mathcal{V}^o$. Preparing training data by collecting videos using existing tools/methods to segment foreground for them is also an alternative. However, this way is not recommended since the performance of the foreground extractor may cause occlusion effect and make the accuracy of importance region unstable.

To produce final resizing frame $\mathcal{R}$ after sampling d-$\mathcal{V}$ to input frame $\mathcal{V}$, we reconstruct the content in a window ($\mathbf{W}$) of $\big[(\frac{S_w}{2}- \frac{T_w}{2}) : (\frac{S_w}{2}+\frac{T_w}{2})\big]$. Since our RETVI is trained to play with arbitrary ratios while controlling several aspects of video quality (such as global content, critical region content preservation, temporal coherence, \textit{
etc.}), the content after sampling may not always fit with the window $\mathbf{W}$. More specifically, when resizing a frame of width $S_w$ to a target width $T_w$, we try to estimate matrix $\mathbf{H}$ such that the total movement of pixels is as tight as possible to $T_w$. However, due to the trace-off in our afore discussion, the total movement would be lightly greater than $T_w$ in some cases. As a result, cropping may occur. See detail visualization in the supplementary file.

\input{frames}

\input{Result_8}

\subsection{Our Results and Discussion}
To demonstrate the abilities of our method, we test it on various video contents. Fig.\ref{fig_frames} exhibits the results of three typical showcases in our experiments, each video has different attributes. To be specific, \textcolor{black}{the video \say{\textit{Dancing}} is with people, video \say{\textit{Talkshow}} is with a single object and text attributes, and the video \say{\textit{Packing gift}} is with multiple objects distributed in entire frame.} Among the three videos, the video \say{\textit{Dancing}} is a camera-move-video. Readers are encouraged to explore our project website (\href{http://graphics.csie.ncku.edu.tw/RETVI}{http://graphics.csie.ncku.edu.tw/RETVI}) to see more results. We can summarize the aspects that enable our results to advance previous works as follows.

We can alleviate the shrinking phenomenon of important objects when reducing sizes effectively. This aspect is gained by benefiting from the CFA module and the critical region loss ($\mathcal{L}_{cri}$). We observe that this phenomenon often involves reducing videos to a considerably smaller size, \textit{i.e.}, less than $0.6$ of width. In the demonstration of Fig.\ref{fig_frames}, we put the results generated by linear scaling to facilitate inferring the quality of our results. It can be seen the dancers (video \say{Dancing}), the lady (video \say{Talkshow}), and the gift-box shape (video \say{Packing gift}) are preserved quite well without shrinking artifacts. \textcolor{black}{It is obvious to observe the video \say{Talkshow}, the shape of the lady is shrunk and the text is distorted when the video is reduced to 0.3 of width by linear scaling. However, they still appear in a visually pleasing manner in our results.}

Being tolerated of various video contents and arbitrary aspect ratios without suffering from distortion could be an advantage of our work. As can be seen in Fig. \ref{fig_frames}, the important contents in three sample videos are not damaged when the resizing ratios varies from 0.9 to 0.3. Particularly in the video \say{Packing gift}, it is such a challenging video since the important contents are dense and distributed in entire frame. Thanks to the performance of our CFA module, our network can fully understand such dense content. Besides, with the aid from global integrity loss $\mathcal{L}_{glo}$, all information of this challenging video is preserved harmoniously and integratively with the source. 

Being extendable to image retargeting is a plus of our proposed system. Several deep learning-based algorithms have been investigated to resize image recently. \textcolor{black}{Detail comparisons with these methods are presented in the supplementary file}. Here, we exhibit the \textcolor{black}{typical} cases mentioned as the challenge in this research domain to demonstrate the capability of our RETVI. The research by \citet{tang2019image} concludes that not all images can be equally resized. This property is defined via a retargetability score, ranged from 0 to 1. They show that the images with low retargetability scores are difficult to preserve visually and semantically important content after resizing. For this challenge, we test our RETVI on the images with different retargetability levels. \textcolor{black}{Fig.\ref{fig_result_8} presents two samples, one is with medium level (score: 0.57), and the other is with low retargetability (score: 0.1).} As shown in the figure, in image (a), the compared method F-MultiOp generates result with serious cropping (\textit{i.e.}, the hand and the mug). In image (b), the shape of the main object (\textit{i.e.}, the clock) is distorted by deformation in AAD \citep{10_panozzo2012robust}. In contrast, these phenomena do not occur in our results. Our RETVI produces plausible results in both cases. The results of this experiment demonstrate that our method is potentially used for retargeting images with various image content.

\input{Compare_Timing}

Finally, running time could be an essential advantage of our method over prior work. Table \ref{compare_timing} reports \textcolor{black}{the running time (excluding video reading/writing) that our RETVI competes} with a conventional video resizing system \citep{lin2013content}. We run two methodologies on the same PC configuration, \textcolor{black}{as reported in Sec.4.1}. An input video consisting of 144 frames in size of $1280 \times 720$ is retargeted to $50\%$ of the width in this experiment. As can be seen, in \citet{lin2013content}, a time-consuming process falls in generating the saliency map and segmentation, approximately more than 1 minute. And the optimization in the retargeting phase spends about $1.32$ seconds. As a result, this conventional system spends minutes to retarget a frame. Meanwhile, our proposed method is an end-to-end system, which only needs approximately 2.3 seconds to resize a video with 144 frames (\textit{i.e.}, 0.016 seconds per frame). \textcolor{black}{It's worth noting that it could be unfair when comparing the running time between a conventional system versus a deep learning-based one since our RETVI also uses significant cost for training process. Nonetheless, the time cost for training our RETVI is 10 hours, which is such a relatively normal training time for a deep learning model. The workflow and cost make the system \citep{lin2013content} cumbersome to be incorporated in demanding services/devices. On contrary, the advantage of running time of RETVI once trained reveals that it is feasible to embed our scheme into a resizing application/service. }
\input{Result_Adobe}
\input{Result_Alpha}

\noindent\textbf{Potential application. }With the abilities of our method in the above discussion, we can apply it to resize videos to make them perform well in the 9:16 aspect ratio. This aspect ratio became popular when smartphones were created with video capabilities. For example, the optimal measure for an Instagram story is 1080px by 1920px, which means its ratio is 9:16. The same goes for other popular social platforms, e.g., Facebook Reels, Youtube Short videos, Tiktok, Instagram story, \textit{etc}. Several existing commercial applications, \textcolor{black}{Adobe Express} \citep{AdobeExpress}, Veed \citep{Veed}, and Flexier \citep{Flexier} have been providing streamers with a function to edit the ratio of their videos before uploading. Nevertheless, these applications share an identical technique, \textit{i.e.}, a window of size 9:16 is allocated at the middle of video and they manually crop two sides of the video. Therefore, they can keep the ratio of object in the same as in the input video. Yet, in the cases that videos with multiple objects, retaining the important objects in a 9:16 ratio is challenging. We provide a showcase in Fig. \ref{fig_result_adobe}, in which we \textcolor{black}{compare} with \textcolor{black}{Adobe Express}. As can be seen, our system does not crop out the tasteful content in this example. The shape of objects may be smaller than those by \textcolor{black}{Adobe Express}, but it can capture the tasteful moment on each frame and make it visually pleasing. The visual video can be seen here \footnote{\href{http://graphics.csie.ncku.edu.tw/RETVI/CompareAdobe.mp4}{http://graphics.csie.ncku.edu.tw/RETVI/CompareAdobe.mp4}}.

To be specific, in order to change a video to a tall one, we apply a parameter to control the movement weight in equation (6). As a result, this equation becomes:
\textcolor{black}{
\begin{equation}
        \begin{cases}
        \mathbf{H}^{ij}_x = e^s_x \times \Re \times \vartheta \\
        \mathbf{H}^{ij}_y = e^s_y \times \Re \times \vartheta \\
    \end{cases}
\end{equation}}here, $\vartheta$ is a constant greater than 1. This parameter can be adjusted by users such that the preserved content in the tall video can catch the user's expectation. This extension enables users to have more predictable results when editing videos with multiple objects to 9:16 aspect ratio. We provide the results generated by different values of $\vartheta$ in Fig.\ref{fig_result_alpha}. As can be seen, varying value of $\vartheta$ yields different retargeting results. \textcolor{black}{Since the estimated values in matrix $\mathbf{H}$ is capable to avoid distortion, a constant $\vartheta$ produces linear increasement}. The bigger $\vartheta$ can retain object in the tight ratio with those in the input video, but it may not retain much content as the smaller $\vartheta$. Depending on what users interest to preserve, $\vartheta$ can be adjusted to achieve the expected results.

\subsection{Evaluation metrics}
To evaluate the performance of the proposed method, we use four metrics. First, \textcolor{black}{we estimate the distortion degree of retargeted videos}. Second, we measure the stability of the retargeted videos. Third, we use an image quality measurement to evaluate the performance of our method on a benchmark image retargeting dataset. And for the last metric, we base on human visual perception. In this evaluation session, we use a set of nine videos and an image dataset. We collect videos data from Youtube such that the videos have diverse content, e.g., single-moving objects, multiple-moving objects, complex backgrounds, or important content distributed in the entire frame. Four of them are the camera-move-videos, and the remainders are object-move-videos. Since few methods focus on retargeting videos, particularly deep learning-based approach, the competitor we quantitatively compare is a typical video resizing system \citet{lin2013content}. Moreover, the source code of this work is provided by the authors, hence it is reliable to use and fair for comparison/evaluation. The image dataset we use in our assessment is RetargetMe \citep{rubinstein2010comparative}, which consists of 80 images. We evaluate the image data in comparison with four recent SOTA image resizing methodologies.

\subsubsection*{\textbf{E.1.} Video \textcolor{black}{quality} measurement}
As the ground truth for video retargeting is not available, it is challenging to define the quality of videos after retargeting process. Therefore, in this regards, we elaborate as follows.

We first adopt the bidirectional similarity measure \citep{simakov2008summarizing} to evaluate the quality on single video frame. \citet{simakov2008summarizing} propose this measurement to describe the coherence and completeness between input and output images \citep{liang2016objective}. It is widely used for quantitative analysis retargeting results in several works. Given a video with $n$ frames, we have two sets: \textcolor{black}{a set of }the source video frames $\mathcal{V}^s = \{F^s_1, \dots, F^s_n\}$ and the other is those retargeted by a certain video retargeting method $\mathcal{V}^t = \{F^t_1, \dots, F^t_n\}$. On each pair ($F^s_k, F^t_k$), the error of $F^t_k$ over $F^s_k$ is expressed as:
\begin{equation}  
    E_k(F^s_k, F^t_k) = {{1\over N}\bigg(\sum\limits_{p\subset F^s_k}{\min\limits_{q\subset F^t_k}}\delta(p,q)+\sum\limits_{q\subset F^t_k}{\min\limits_{p\subset F^s_k}}\delta(q,p)}\bigg),
\end{equation} where $k \in [1 \dots n]$, $N$ is the number of patches \textcolor{black}{on $F^s_k$ and $F^t_k$} ; $\delta(.)$ is defined by sum of squared distance of two patches. Afterwards, we calculate the mean ($\mathcal{M}_{E}$) of $E_k$ to define the error degree of a retargeted video. The lower is better.

\subsubsection*{\textbf{E.2.} Video stability measurement}
To measure the stability of resized videos without annotated data, we \textcolor{black}{estimate} the differences of adjacent frames in the retargeted video and compare them against those in the source one. This concept is used in various video generation applications, \textit{e.g.,} video stylization \citep{deng2021arbitrary, lestructure}, video resequencing \citep{10018537}. The calculation is based on the fact that the source videos are temporally coherent, and the retargeted results are rendered from the same frame set with them but in different ratios. We, therefore, treat the value in the source as the ground truth to judge the stability degree of the resulting videos. Accordingly, a generated video with a score that is tightly asymptotic to the ground truth would be in good stability.

Given a video with frame set $\mathcal{V} = \{F_t\}$, $t\in [1 \dots n]$, $n$ is the total frames of $\mathcal{V}$, the difference degree of two adjacent frames is formulated as:
\begin{equation}
    \textcolor{black}{\mathcal{D}_{t \rightarrow t-1} = \frac{1}{100} \times \frac{\big |F_t - F_{t-1}\big|}{H \times W},}
\end{equation}
\textcolor{black}{where $H$ and $W$ are the height and width of frames in set $\mathcal{V}$. $\big |F_t - F_{t-1}\big|$ returns a residual of two frames, which can be seen in the supplementary.}
Accordingly, the stability of the video is:
\begin{equation}
    STB = \frac{1}{n}\sum^n\limits_{t=1}\mathcal{D}_{t \rightarrow t-1}.
\end{equation}For this metric, the lower score represents a better stability. 

In Table \ref{table_compare_Lin}, we analyze the quantitative evaluation and comparison using the above two metrics, $\mathcal{M}_E$ and $STB$. With nine videos we prepared, we used \citet{lin2013content}'s method and our RETVI to generate results. As can be seen, our method achieves a lower distortion on average compared to Lin's. The highest $\mathcal{M}_E$ in our results is 3.17; two videos have relatively low distortion (\textit{i.e.}, video \#4 and video \#9). Inferring to Lin's results, the distortion level in theirs is higher than ours in most cases. The lowest $\mathcal{M}_E$ of their results is reported at 2.97, which is higher than our average score. This analysis reveals that the retargeting results generated by our system are less distorted than \citet{lin2013content}. In terms of stability, Lin’s method and our RETVI have comparable scores. However, we can see that their values of $STB$ of the camera-move-videos are relatively higher than those of the reverse cases. The analysis results reveal that our model is stable when working on diverse input videos.

\input{Charts}
\subsubsection*{\textbf{E.3.} Image quality measurement}
Although our focus is not retargeting images, as mentioned above, our method effectively applies to single images. To quantitatively evaluate this performance, we adopt the Aspect ratio similarity (ARS) \citep{zhang2016aspect}. This algorithm evaluates the visual quality of retargeted images by exploiting the local block changes with a visual importance pooling strategy. The ARS score represents the geometric changes during retargeting \citep{zhang2016aspect}. For this metric, all of the images in the RetargetMe are used. We compare the results of these images generated by our method against a typical conventional method, multi-operator \citep{15_rubinstein2009multi}, and three deep learning-based techniques \citep{cho2017weakly, tan2019cycle, zhou2020weakly}. The results of multi-operator \citep{15_rubinstein2009multi} are encompassed in the RetargetMe dataset, while those of WSSDCNN \citep{cho2017weakly}, Cycle-IR \citep{tan2019cycle}, and SAMIR \citep{zhou2020weakly} are generated from the source code released by the authors.

\input{Compare_vs_Lin}

The quantitative comparison of the ARS score is shown in Fig. \ref{fig_Charts}(c). As can be seen, our RETVI achieves a higher score comparing all the competitors. Although the deviation is not significant, it is sufficient to demonstrate our method’s performance. It is because the RetargetMe dataset encompasses challenging images. A good performance on this dataset could be assessed as a reliable method. 

\subsubsection*{\textbf{E.4.} User study}
We further estimate the performance of our RETVI based on human visual perception. We conduct a user study on two groups. In one group (G-1), we recruit 13 users (nine of them with graphics-related backgrounds). For the other (G-2), we invite nine users who are either tiktokers or senior users on the existing social platforms. In G-1, we use nine videos mentioned in evaluations E.1, E.2 to generate results retargeting to a half size of width by our RETVI and \citet{lin2013content}. In G-2, we prepare five videos (two with a single object and three with multiple objects) and resize them to the 9:16 aspect ratio by \textcolor{black}{Adobe Express} \citep{AdobeExpress} and our RETVI. In both groups, participants are shown two retargeted videos at one time and asked to choose the one they prefer. We receive 117 responses on G-1 and 45 responses on G-2. We then compute the percentage of votes for each video.

Fig. \ref{fig_Charts}(a)-(b) shows the statistics of users' preferences. It can be seen that our method receives majority votes from the participants in G-1 and G-2 alike. There are two cases in G-1 (e.g., vid\#2 and \#6), in which users judge Lin's results are better than ours. However, the difference is not significant (0.08 and 0.14, respectively), and the score of our RETVI is relatively higher in the remaining seven cases (77.78\% in total showcases). The data in G-2 reveals that users prefer the tall videos by \textcolor{black}{Adobe Express} to ours if the videos have a single object. Yet, our RETVI wins \textcolor{black}{Adobe Express} in cases with multiple objects.
\input{Compare_3}
\input{Compare_4}

\subsection{Visual comparisons to prior work}
Fig. \ref{fig_compare_3} visualizes the comparison between our results and \citet{lin2013content}'s. The visual results reveal that Lin's results suffer from some shortcomings. First, several noticeable artifacts occur in the main object and line structure of the background region (highlighted by yellow rectangles). The reason is that \textcolor{black}{their scheme} relies on a saliency map \citep{goferman2011context} to estimate the movement weight of quad vertices. As a result, if the content frame is dense, the saliency method may not be tolerated to analyze the frame effectively. For example, the shape of the lady in frame \#30 and frame \#87 is plausible, but the door is shrunk significantly. It is even smaller than linear scaling. Besides, in frame \#199, the line structure of the door appears quite well, but the head of the lady is distorted. \textcolor{black}{In contrast,} our method produces more visually pleasing results without these distortions. Second, the generated video by \citet{lin2013content} in this case still has the noticeable flicking artifact. See the video here\footnote{\href{http://graphics.csie.ncku.edu.tw/RETVI/CompareLin.mp4}{http://graphics.csie.ncku.edu.tw/RETVI/CompareLin.mp4}} for better visualization.

In Fig. \ref{fig_compare_4}, we provide a visual comparison with \citet{lee2020object}, an object detection approach for retargeting video. As can be observed, there is no artifact or distortion in their results and ours alike. The noticeable point here is the ratio of objects in the retargeted frames. With the goal that the contents inside the bounding boxes must remain intact in the retargeted frame, \citet{lee2020object} do a good job, but their appearance is in a close ratio with those in the input frames. For example, visually inspecting this figure, the background contents (e.g., the door on the left, the wall on the right, or the plants) are scaled up when the width increases in this experiment. However, the shape of the lady is hard to recognize the changes in the ratio over the source frame. These may reduce the harmonization of the results. In contrast, our method does not suffer as they do. Our results are more harmonious in both background contents and important objects. The visual video can be seen here\footnote{\href{http://graphics.csie.ncku.edu.tw/RETVI/CompareLee.mp4}{http://graphics.csie.ncku.edu.tw/RETVI/CompareLee.mp4}}.

Apart from the above comparisons with video retargeting methodologies, we further discuss the ability of our method via visual comparison with four recent state-of-the-art image retargeting systems, WSSDCNN \citep{cho2017weakly}, SAMIR \citep{zhou2020weakly}, grid encoding model \citep{kim2018quad} and Cycle-IR \citep{tan2019cycle}. Detail of discussion and visualization on this comparison is presented in the supplementary file.

\subsection{Ablation Study}
\vspace{-0.25cm}
\subsubsection*{A.1. Verify the effectiveness of CFA module}
The CFA module is proposed to alternate the time-consuming preprocessing phase in such a conventional video retargeting system. The CFA is structured as an improvement from a plain UNet design. Our proposed CFA can capture more fine-grained feature, which is an essential key for our retargeting system to tolerate with diverse content videos. We demonstrate the effectiveness of CFA module by removing it from our training and train our RETVI with the plain UNet \citep{ronneberger2015u}. We show these ablation analyses in Fig. \ref{fig_visualize_Grad}, in which the visualization of Grad-CAMs \citep{selvaraju2017grad} and energy maps are obtained from our network with and without the CFA module. The results reveal that with the CFA module, our network has much larger attended regions. This enables our RETVI to play with more complex video contents without suffering from distortion and linear-like appearance. 
\input{Visualize_Grad}

\subsubsection*{A.2. Study on the impact of loss functions}
Performance of our RETVI is affected by the optimization of the loss functions. Here, we discuss the impact of each component we configured in the objective function. To verify this, we remove in turn each loss function in Eq. (8) from our training. The ablated results are presented in Fig. \ref{fig_ablated_result}. It is obvious that without each component in the total loss function, it is challenging to obtain ideal retargeting results. To be specific, we can see that without the guidance of $\mathcal{L}_{cri}$, the important region in the frame is damaged. On the one hand, the object in the foreground is shrunk linearly, on the other hand, the background captured as critical region by our CFA is distorted. We also discuss the effect of the global loss $\mathcal{L}_{glo}$ in Fig.\ref{fig_ablated_result}(b). Without $\mathcal{L}_{glo}$, we fail to preserve the initiative of the content frame. \textcolor{black}{Meanwhile, removing $\mathcal{L}_{fid}$ could yield resizing form without damaging the content but blurry effect is a negative side}. The full configuration facilitates our model producing visual pleasing result. \textcolor{black}{Also, the absence of $\mathcal{L}_{tem}$ causes serious flickering artifact in the generated videos. The aid of this loss in our objective serves resulting videos in good stability. See the ablated result of $\mathcal{L}_{tem}$ here\footnote{\href{http://graphics.csie.ncku.edu.tw/RETVI/TemporalLoss.mp4}{http://graphics.csie.ncku.edu.tw/RETVI/TemporalLoss.mp4}}.} 

\input{Ablated_Result}

\textbf{Limitation. }In the cases that the input videos are the landscape scenes or too dense, our method may not perform well. The failure phenomenon here is that the retargeted videos are quite similar to linear scale. We note here that distortion does not occur in these cases. It is because our ADE module fails to differentiate which object/region is important. Thus, the estimated energy of pixels is relatively identical. Another limitation falls into the failure of the loss $\mathcal{L}_{glo}$. Since we rely on performance of ResNet to define the integrity of estimated frame versus the input one, the failure of ResNet leads to $\mathcal{L}_{glo}$ to be disabled. The $\mathcal{L}_{glo}$ performs well in most of the cases, but it is not guaranteed to be stable in the videos with multiple objects distributed in entire frame. \textcolor{black}{Another limitation is that if the main objects located in the leftmost of rightmost, the resized results may not appealing due to the cropping effect as we discussed in Sec.4.1. An example in this case is visualized in the supplementary file. }

\section{Conclusion}
In this paper, we propose a new RETVI framework for retargeting videos. With two modules configured in our method, our RETVI presents high performance in handling videos with diverse contents and produces visually pleasing results when retargeting to arbitrary aspect ratios. The analysis and experimental results demonstrate that our method substantially advances prior works. With the fast running time of our end-to-end RETVI, our system is potentially embedded into a video resizing application/service. \textcolor{black}{We perceive that our system can bypass the computational bottlenecks in conventional methods. And it is potential to extend for stereo image/video retargeting}. For the shortcoming we discussed, we plan to investigate techniques that configure the loss function to be independent from the existing feature extractor. \textcolor{black}{In terms of cropping effect, a possible way can improve is automatically define physical region of the important content. This knowledge could serve us to shift the rendering window more appropriately}. Besides, developing a text-driven framework to consider semantic issue in the retargeting videos \textcolor{black}{and investigating technique to retarget videos with enlarging and reducing two dimensions simultaneously are also a possibly extension in our near future}. This could be a potential way to visualize users' expectation in such a video retargeting system.

\bibliographystyle{abbrv}
\bibliography{references}
\end{document}

%% file: Timing_images.tex
\begin{figure}
  \centering
  \includegraphics[width=0.65\linewidth]{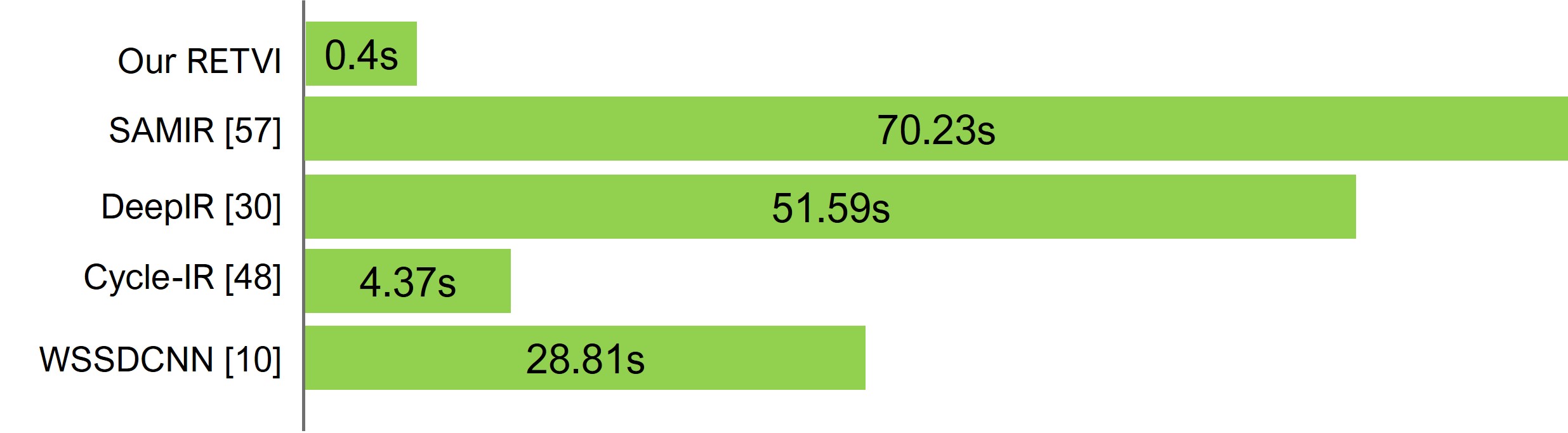}
  \caption{\small \textcolor{black}{Comparison on running time (second) of the existing SOTA image retargeting methods and our RETVI.}}
  \label{fig_timing_images}
\end{figure}

%% file: Net.tex
\begin{figure*}
  \centering
  \includegraphics[width=\linewidth]{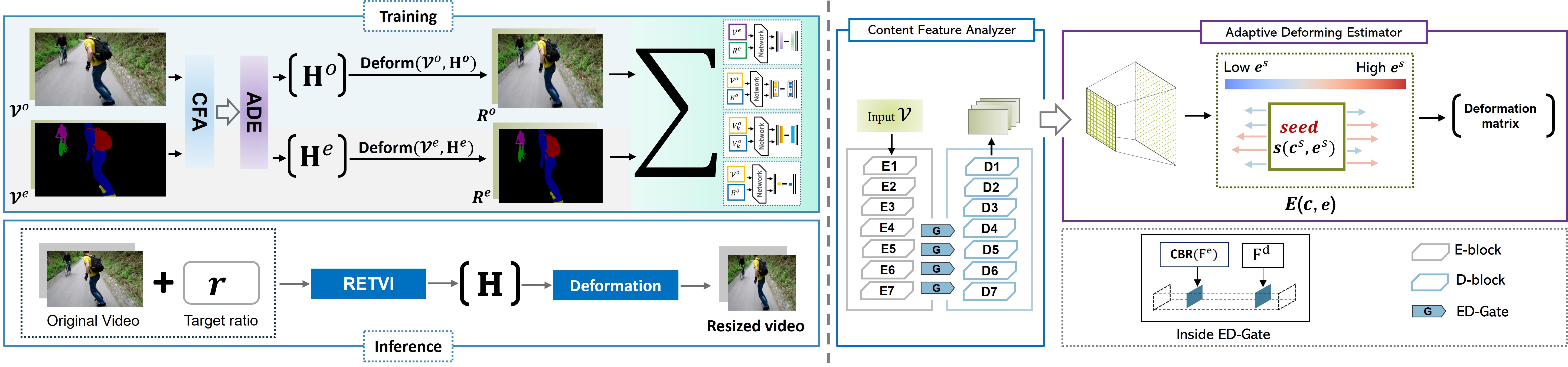}
  \caption{\small \textcolor{black}{Left: the workflow of our RETVI in training and inference; right: visualization of the CFA and ADE modules.}}
  \label{fig_Net}
\end{figure*}

%% file: workflow.tex
\begin{figure}
  \centering
  \includegraphics[width=0.65\linewidth]{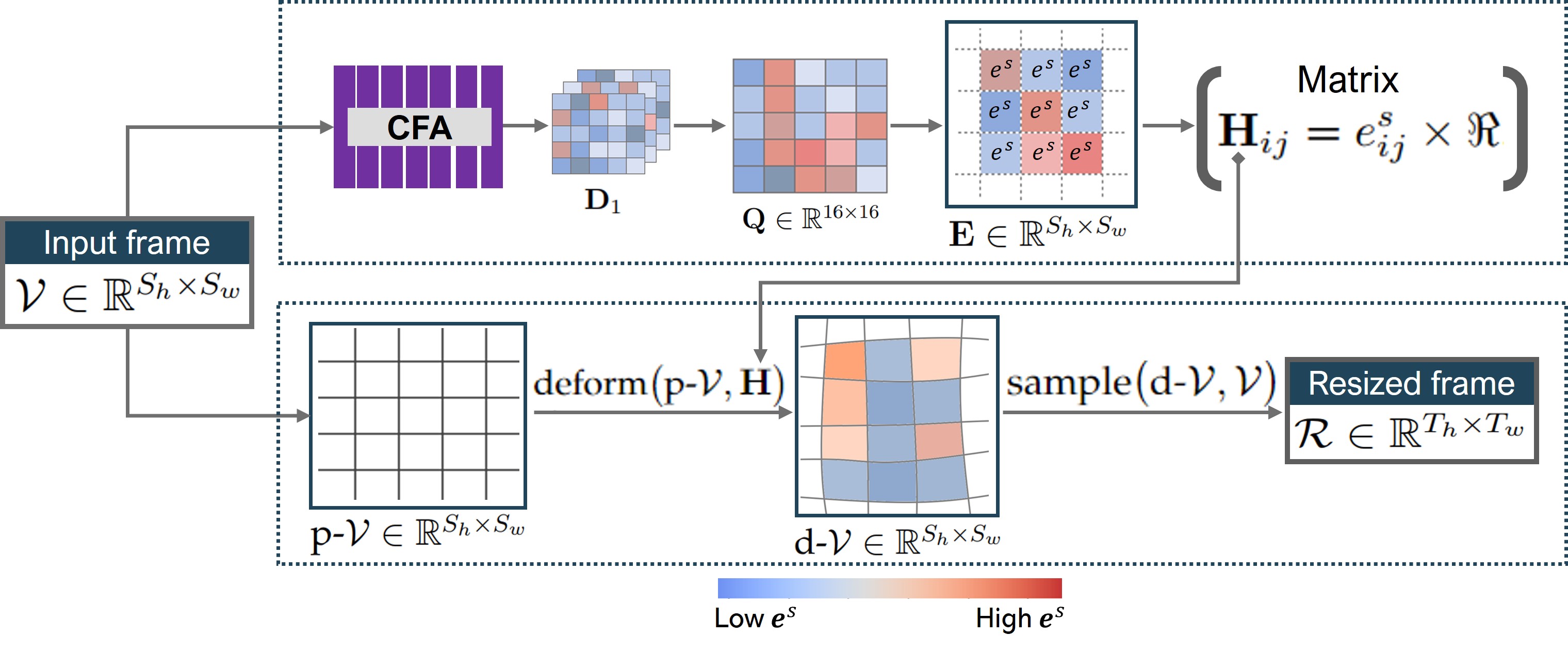}
  \caption{\small The workflow of our resizing strategy}
  \label{fig_workflow}
\end{figure}

%% file: frames.tex
\begin{figure*}
  \centering
  \includegraphics[width=\linewidth]{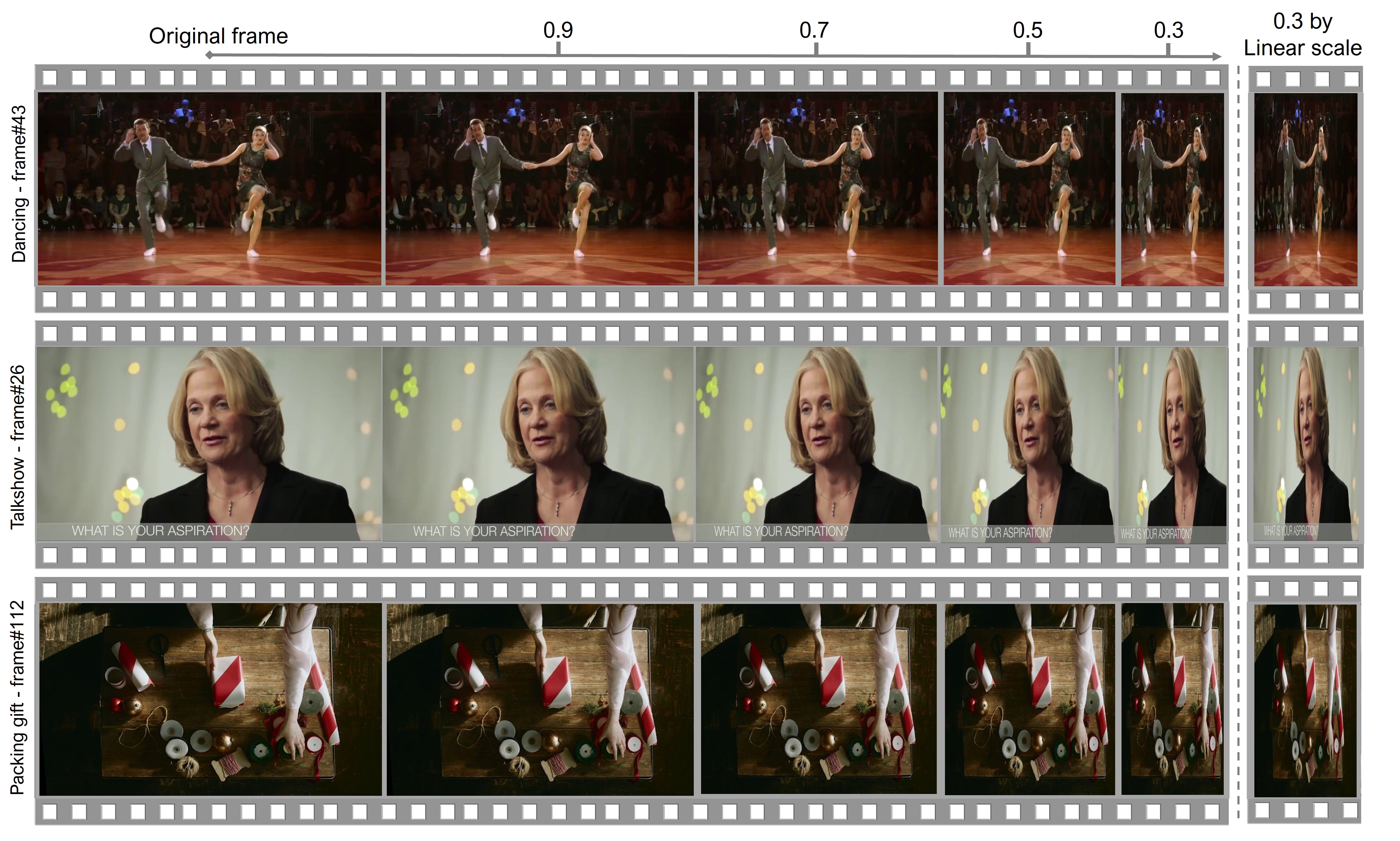}
  \caption{\small Performance of our method on different video contents by retargeting to ratio from 0.9 to 0.3 of width.}
  \label{fig_frames}
\end{figure*}

%% file: Result_8.tex
\begin{figure*}
  \centering
  \includegraphics[width=0.95\linewidth]{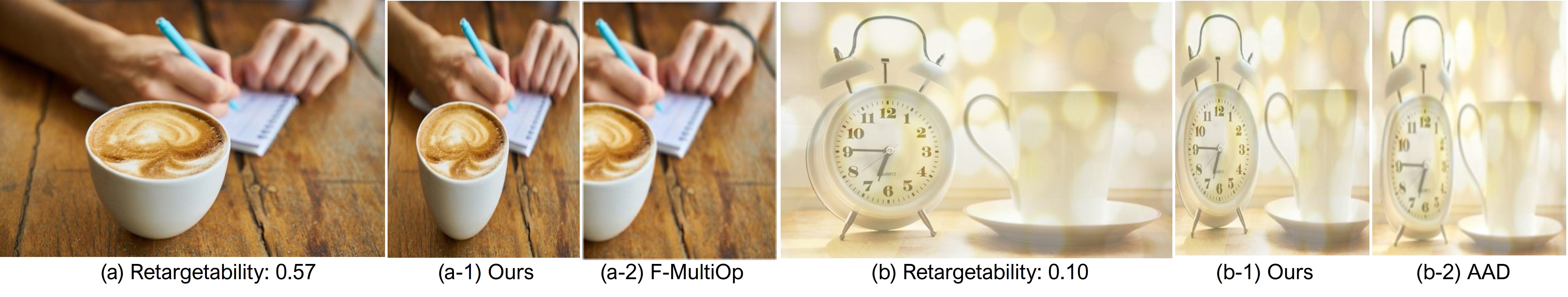}
  \caption{\small Our method challenges on images with different retargetability levels. The input images and results by F-MultiOp and AAD are quoted from the source paper \citep{tang2019image}.}
  \label{fig_result_8}
\end{figure*}

%% file: Compare_Timing.tex
\begin{table}[h!]
  \centering
  \caption{\small Comparison on execution time per frame}
    \begin{tabular}{c c c c}
    \\[0.5ex]
     \Xhline{2\arrayrulewidth} 
    Method & \multicolumn{2}{c}{Pre-processing} & Retargeting \\
    \cmidrule(lr){1-1}\cmidrule(lr){2-3}\cmidrule(lr){4-4}
    Ref. \citep{lin2013content} & \textit{\small Saliency map}-4(s) & \textit{\small Segmentation}-72(s) & 1.32(s) \\
    \cmidrule(lr){1-4}    
    Ours  & No    & No    & 0.016(s)\\
   \Xhline{2\arrayrulewidth} 
    \end{tabular}%
  \label{compare_timing}%
\end{table}%

%% file: Result_Adobe.tex
\begin{figure}
  \centering
  \includegraphics[width=\linewidth]{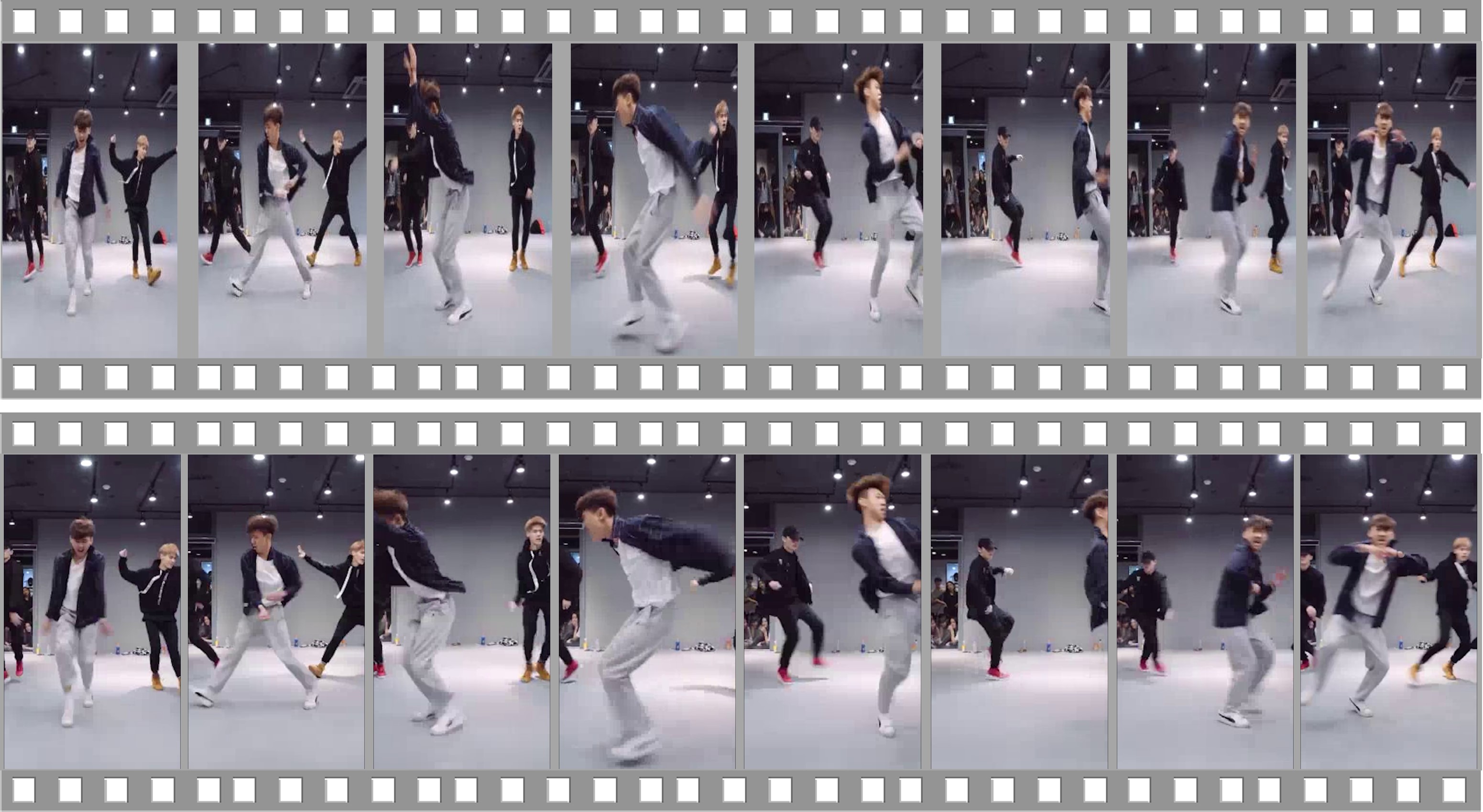}
  \caption{\small Our RETVI (the first row) competes with Adobe Express (the second row) on resizing video to 9:16 aspect ratio. }
  \label{fig_result_adobe}
\end{figure}

%% file: Result_Alpha.tex
\begin{figure}[h!]
  \centering
  \includegraphics[width=\linewidth]{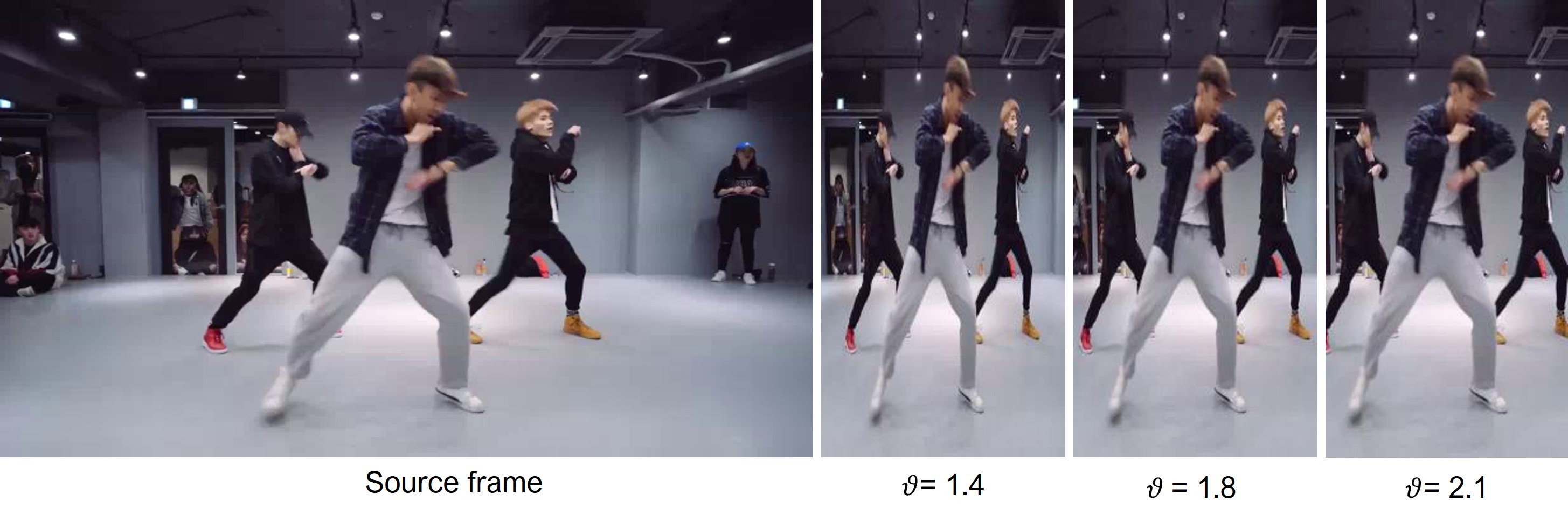}
  \caption{\small Predictable results for tall video.}
  \label{fig_result_alpha}
\end{figure}

%% file: Charts.tex
\begin{figure}
  \centering
  \includegraphics[width=\linewidth]{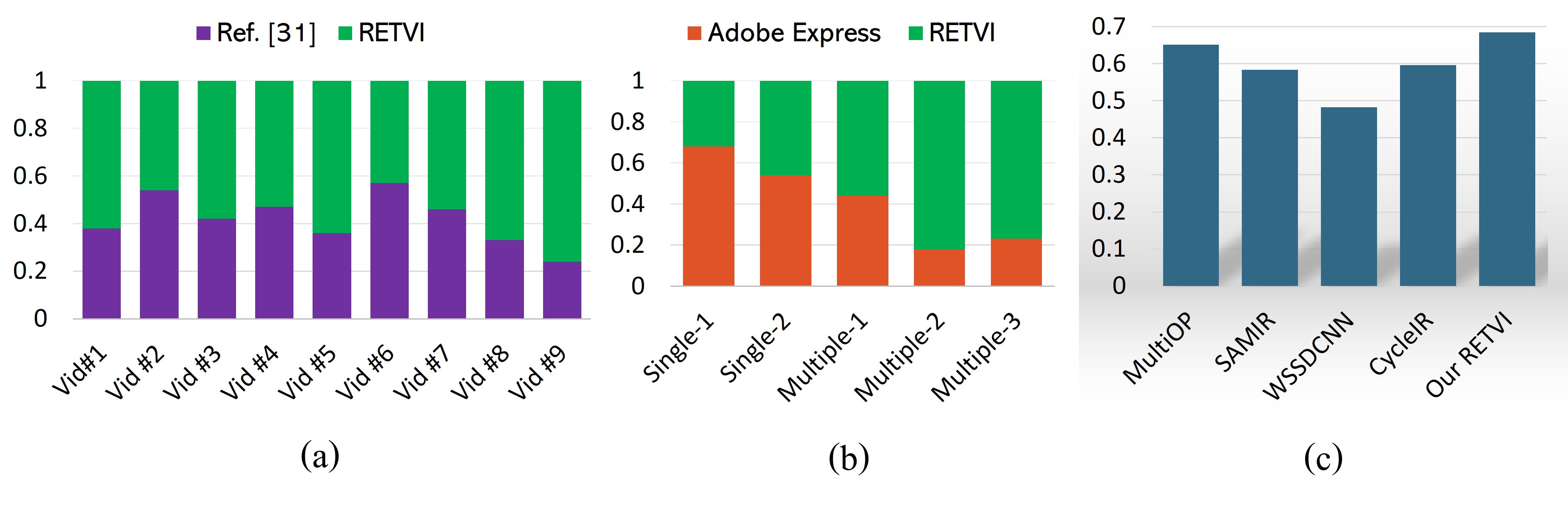}
  \caption{\small Analysis of human perception on group G-1 (a) and  group G-2 (b). (c) is the comparison on ARS score.}
  \label{fig_Charts}
\end{figure}

%% file: Compare_vs_Lin.tex
\begin{table}[h!]
  \centering
  \caption{\small Analysis on the quality of video results - The last four rows are camera-move videos}
    \begin{tabular}{lccccc}
    \\[0.5ex]
    \Xhline{2\arrayrulewidth} 
          Methods& Source video & \multicolumn{2}{c}{\citet{lin2013content}} & \multicolumn{2}{c}{Our method} \\
          \cmidrule(lr){1-1}\cmidrule(lr){2-2} \cmidrule(lr){3-4} \cmidrule(lr){5-6}
          Testing data & $STB\downarrow$   & \textcolor{black}{$\mathcal{M}_{E}\downarrow$}   & $STB\downarrow$ & \textcolor{black}{$\mathcal{M}_{E}\downarrow$}   & $STB\downarrow$ \\
        \cmidrule(lr){1-6}
        1-Talkshow  & 0.057      & \textcolor{black}{3.24}    & 0.062    &
        \textcolor{black}{2.81}    & 0.063  \\ [1ex] 
        2-Packing gift & 0.041       & \textcolor{black}{3.71}    & 0.058    & \textcolor{black}{2.63}    & 0.055  \\[1ex] 
        3-Holding mug & 0.052       & \textcolor{black}{4.15}    & 0.066    & \textcolor{black}{3.16}    & 0.061  \\[1ex] 
        4-Cap movie & 0.063       & \textcolor{black}{2.97}    & 0.071    & \textcolor{black}{1.97}    & 0.06  \\[1ex] 
        5-Shewing & 0.047       & \textcolor{black}{3.62}    & 0.051    & \textcolor{black}{2.84}    & 0.048  \\[1ex] 
        6-Masha & 0.081  & \textcolor{black}{3.52}  & 0.09    & 
        \textcolor{black}{3.17}    & 0.082  \\[1ex] 
        7-Dancing & 0.075      & \textcolor{black}{4.83}    & 0.082    & \textcolor{black}{2.69}    & 0.078  \\[1ex] 
        8-Moana Movie & 0.089      & \textcolor{black}{4.29}    & 0.1    & \textcolor{black}{2.24}    & 0.09  \\[1ex] 
        9-Uptown dance & 0.084      & \textcolor{black}{3.74}    & 0.094    & \textcolor{black}{1.86}    & 0.087  \\[1ex] 
        \cmidrule(lr){1-6}
        \textbf{Average} & 0.065  & 3.78    & 0.74    & 2.59    & 0.069  \\ [1ex]
    \Xhline{2\arrayrulewidth}     
    \end{tabular}%
  \label{table_compare_Lin}%
\end{table}%

%% file: Compare_3.tex
\begin{figure}
  \centering
  \includegraphics[width=\linewidth]{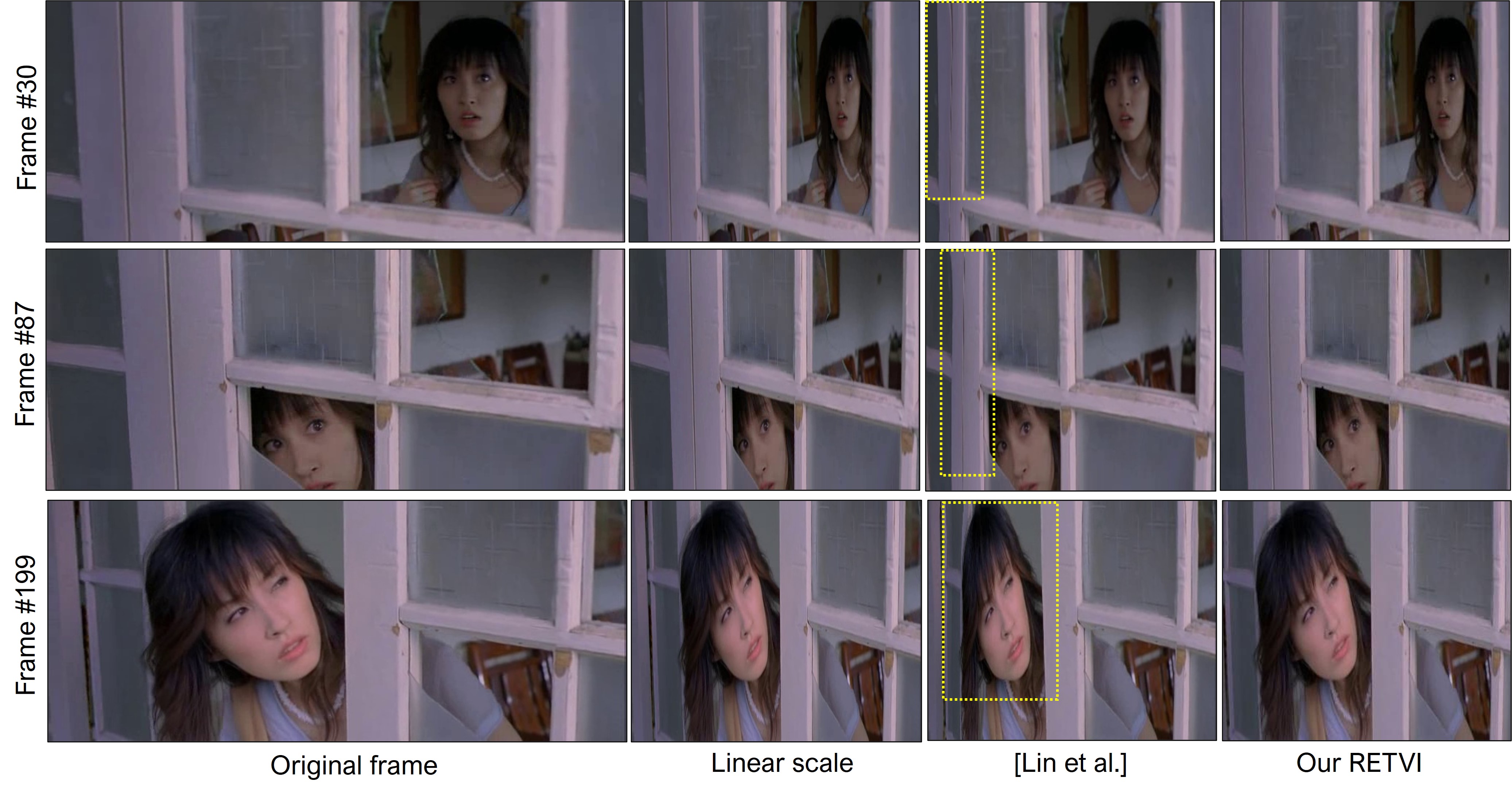}
  \caption{\small Comparison with \citet{lin2013content}}
  \label{fig_compare_3}
\end{figure}

%% file: Compare_4.tex
\begin{figure}[h!]
  \centering
  \includegraphics[width=\linewidth]{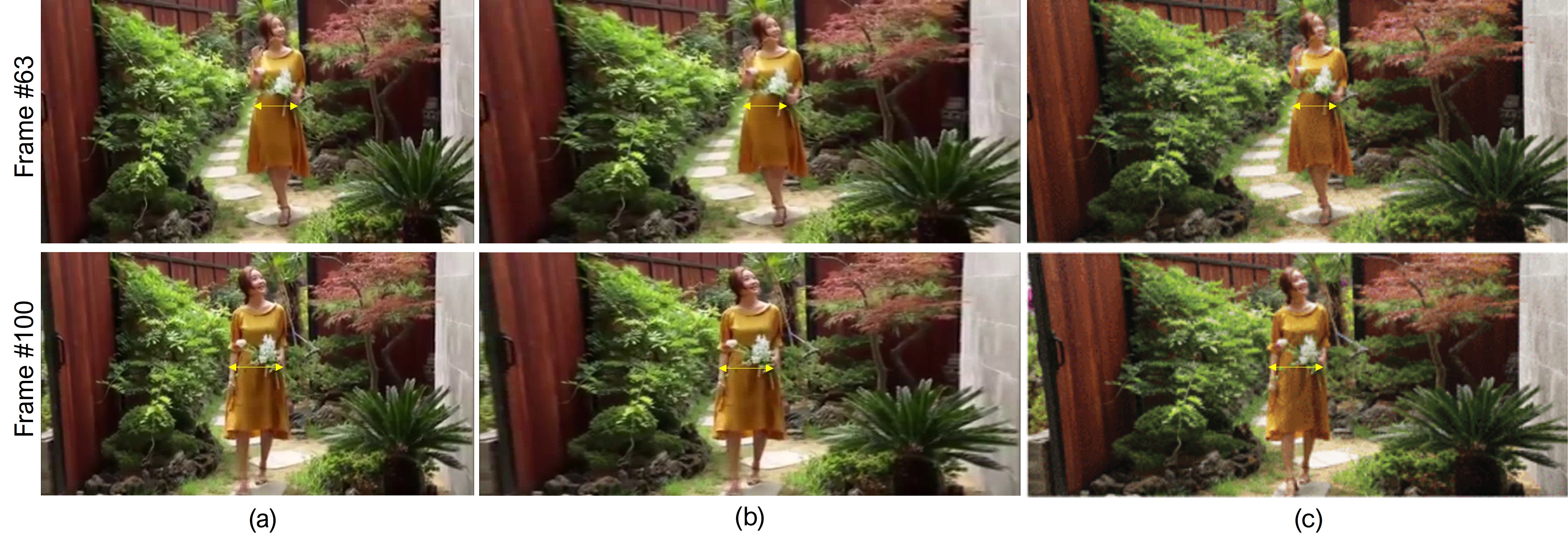}
  \caption{\small Comparison between our results (b) with \citet{lee2020object} (c). Results in (c) are quoted from the source paper \citep{lee2020object}.}
  \label{fig_compare_4}
\end{figure}

%% file: Visualize_Grad.tex
\begin{figure}
  \centering
  \includegraphics[width=\linewidth]{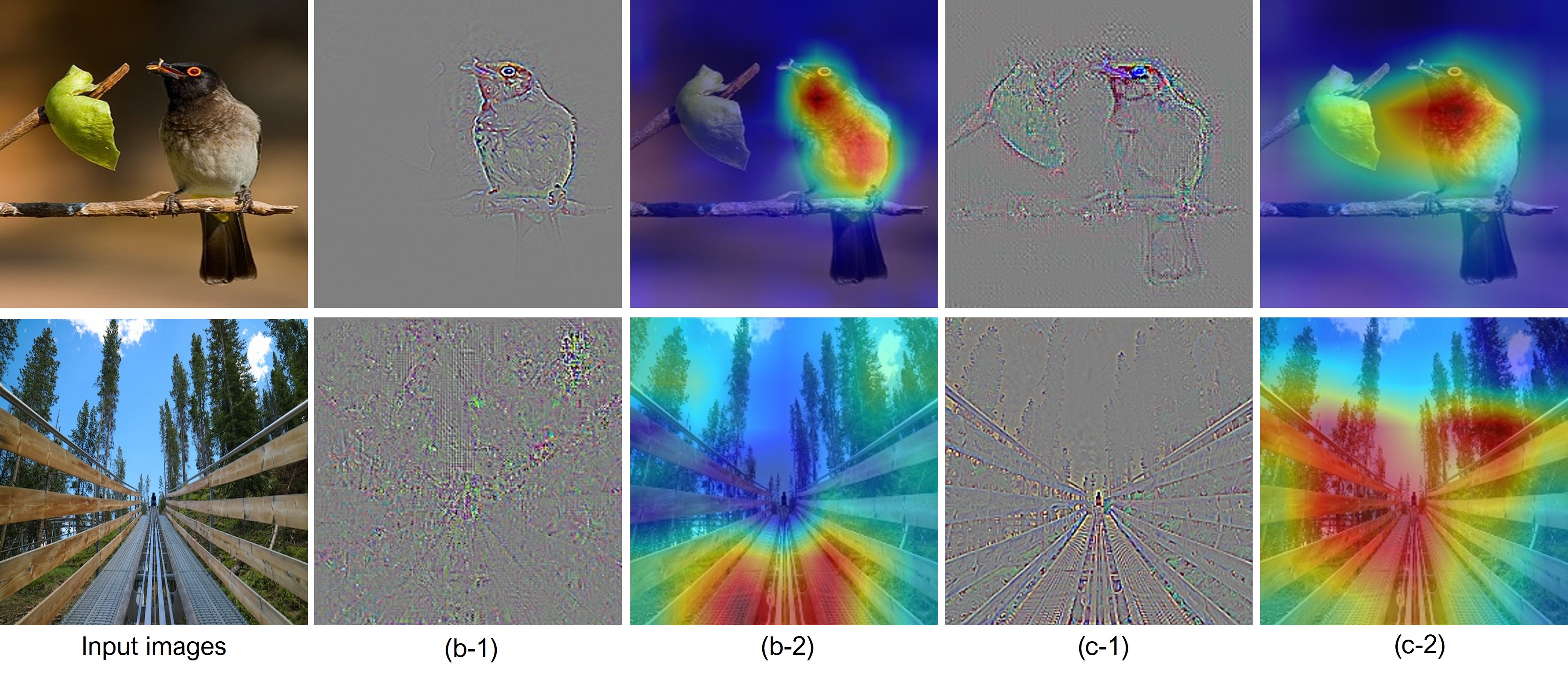}
  \caption{\small The Grad-Cam visualization (b-2, c-2) and energy map (b-1, c-1) of our CFA module (c- columns) versus the standard UNet (b- columns).}
  \label{fig_visualize_Grad}
\end{figure}

%% file: Ablated_Result.tex
\begin{figure}
  \centering
  \includegraphics[width=\linewidth]{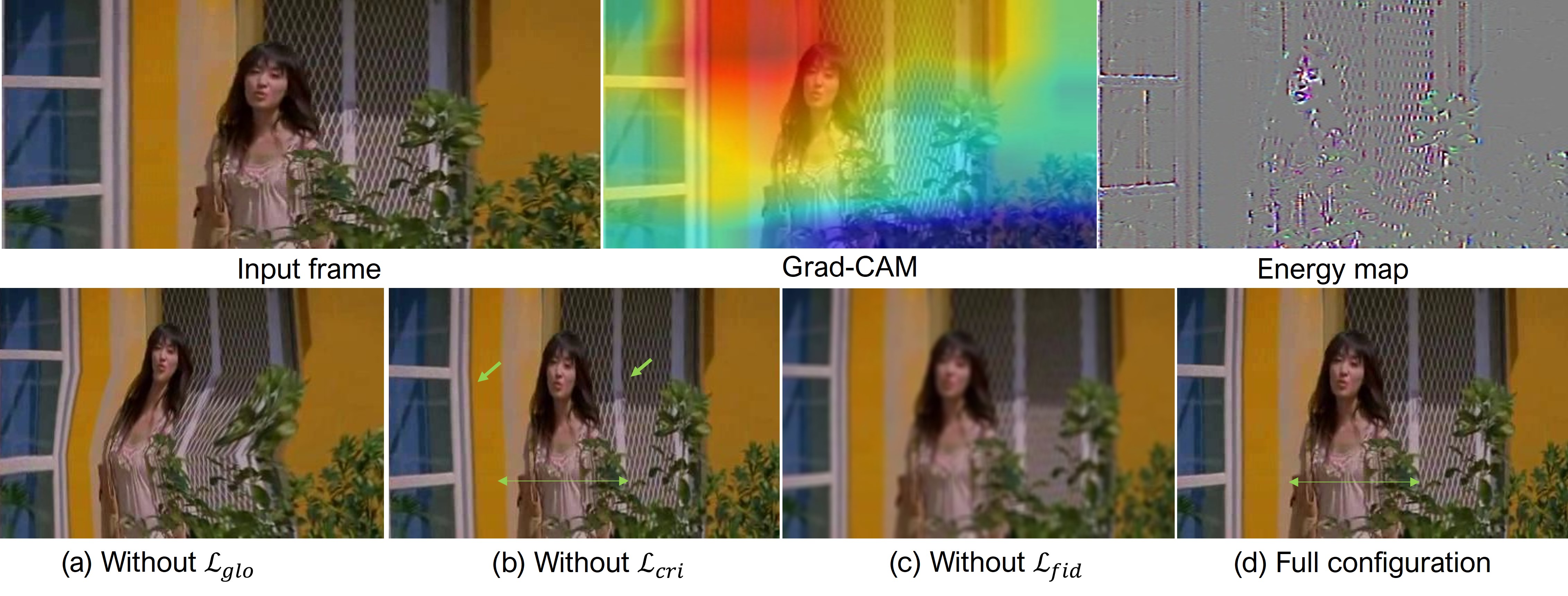}
  \caption{\small \textcolor{black}{Ablated results for loss function. The resized results (second row) in this experiment are resized to ratio 0.7.}}
  \label{fig_ablated_result}
\end{figure}